\documentclass[sigconf]{acmart}

\AtBeginDocument{%
  \providecommand\BibTeX{{%
    \normalfont B\kern-0.5em{\scshape i\kern-0.25em b}\kern-0.8em\TeX}}}

\copyrightyear{2023}
\acmYear{2023}
\setcopyright{rightsretained}
\acmConference[MM '23]{Proceedings of the 31st ACM International Conference on Multimedia}{October 29-November 3, 2023}{Ottawa, ON, Canada}
\acmBooktitle{Proceedings of the 31st ACM International Conference on Multimedia (MM '23), October 29-November 3, 2023, Ottawa, ON, Canada}
\acmDOI{10.1145/3581783.3612426}
\acmISBN{979-8-4007-0108-5/23/10}

\newcommand{\yuyan}[1]{{\color{black} #1}}
\usepackage{rotating}
\usepackage{colortbl}
\usepackage{booktabs}
\usepackage{multirow}
\usepackage{hhline}
\usepackage{paralist}
\usepackage{balance}
\usepackage{enumitem}
\usepackage{subcaption}
\usepackage{graphicx}
\begin{document}

\title[Combating Online Misinformation Videos: Characterization, Detection, and Future Directions]{Combating Online Misinformation Videos: \\ Characterization, Detection, and Future Directions}

\author{Yuyan Bu}
\affiliation{%
  \institution{Institute of Computing Technology, Chinese Academy of Sciences}
  \institution{University of Chinese Academy of Sciences}
  \state{}
  \country{}}
\email{buyuyan22s@ict.ac.cn}

\author{Qiang Sheng}
\affiliation{%
  \institution{Institute of Computing Technology, Chinese Academy of Sciences}
  \state{}
  \country{}}
\email{shengqiang18z@ict.ac.cn}

\author{Juan Cao}
\affiliation{%
  \institution{Institute of Computing Technology, Chinese Academy of Sciences}
  \institution{University of Chinese Academy of Sciences}
  \state{}
  \country{}}
\email{caojuan@ict.ac.cn}

\author{Peng Qi}
\affiliation{%
  \institution{Institute of Computing Technology, Chinese Academy of Sciences}
  \state{}
  \country{}}
\email{pengqi.qp@gmail.com}

\author{Danding Wang}
\affiliation{%
  \institution{Institute of Computing Technology, Chinese Academy of Sciences}
  \state{}
  \country{}}
\email{wangdanding@ict.ac.cn}

\author{Jintao Li}
\affiliation{%
  \institution{Institute of Computing Technology, Chinese Academy of Sciences}
  \state{}
  \country{}}
\email{jtli@ict.ac.cn}

\thanks{The authors are also with Key Lab of Intelligent Information Processing, Institute of Computing Technology, Chinese Academy of Sciences.}

\renewcommand{\shortauthors}{Yuyan Bu, Qiang Sheng, Juan Cao, Peng Qi, Danding Wang, and Jintao Li}

\begin{abstract}
With information consumption via online video streaming becoming increasingly popular, misinformation video poses a new threat to the health of the online information ecosystem. Though previous studies have made much progress in detecting misinformation in text and image formats, video-based misinformation brings new and unique challenges to automatic detection systems: 1) high information heterogeneity brought by various modalities, 2) blurred distinction between misleading video manipulation and nonmalicious artistic video editing, and 3) new patterns of misinformation propagation due to the dominant role of recommendation systems on online video platforms. To facilitate research on this challenging task, we conduct this survey to present advances in misinformation video detection. We first analyze and characterize the misinformation video from three levels including signals, semantics, and intents. Based on the characterization, we systematically review existing works for detection from features of various modalities to techniques for clue integration. We also introduce existing resources including representative datasets and useful tools. Besides summarizing existing studies, we discuss related areas and outline open issues and future directions to encourage and guide more research on misinformation video detection. The corresponding repository is at \url{https://github.com/ICTMCG/Awesome-Misinfo-Video-Detection}.
\end{abstract}


\maketitle

\begin{figure}[t]  
    \centering    
    \includegraphics[width=1\linewidth]{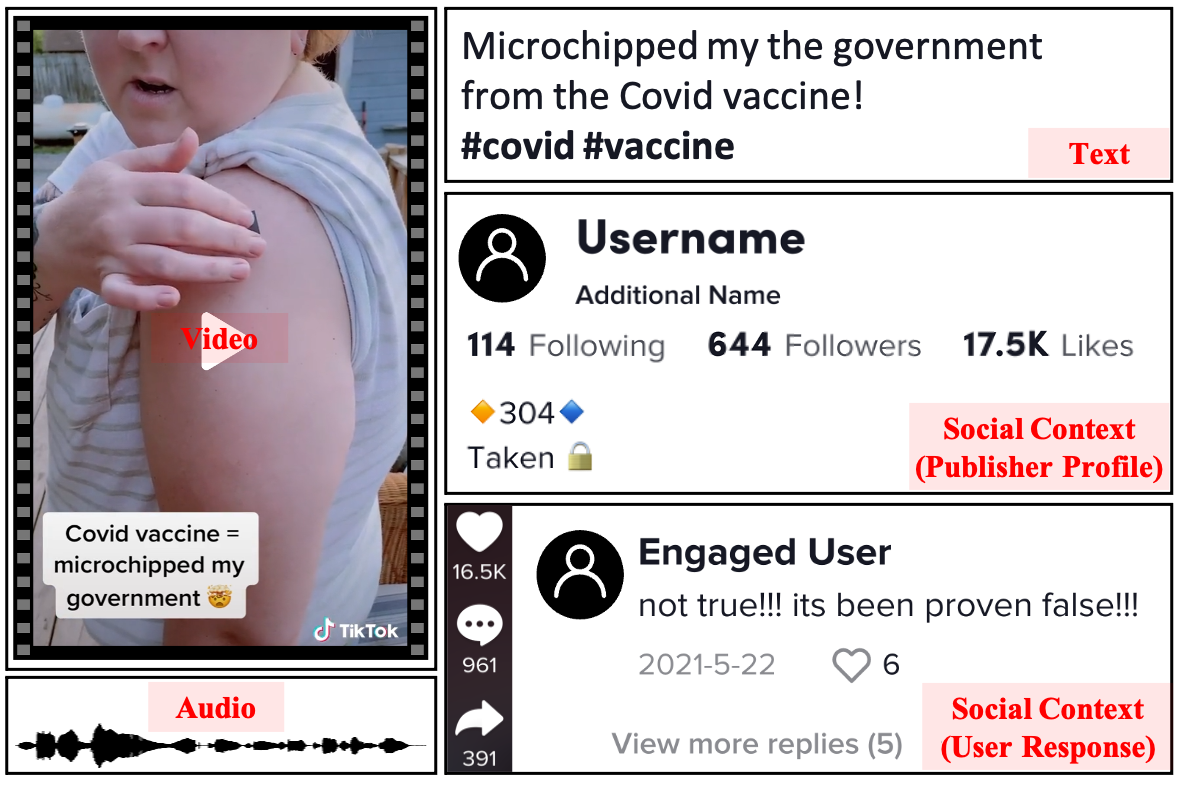}
    \caption{A misinformation video post on TikTok, along with the attached social context information, indicating that COVID-19 vaccines contain microchips. For privacy concerns, we replace the user avatars and names with placeholders.}
    \label{fig:videopost} 
    \vspace{-2em}
\end{figure}

\section{Introduction}
With the prevalence of online video platforms, information consumption via video streaming is becoming increasingly prominent. Popular video-sharing platforms like YouTube and TikTok have attracted billions of monthly active users~\cite{tiktok-youtube-stats}. 
Studies on news consumption show that about a quarter of U.S. adults under 30 regularly get news from these video-sharing platforms~\cite{TikTokNewsConsumption}.

Unfortunately, the massive growth of video news consumption also boosts the rapid spread of misinformation videos, posing an increasingly serious challenge to the online information ecosystem.
For instance, 124 TikTok misinformation videos about COVID-19 vaccines gathered over 20M views and 300K shares~\cite{TiktokSample}.
Fig.~\ref{fig:videopost} shows a misinformation video post about the existence of microchips in COVID-19 vaccines.
Compared with previously studied misinformation in text and image format mostly, video-based misinformation is more likely to mislead the audience and go viral.
Research in the political domain shows that individuals are more likely to believe an event's occurrence when it is presented in video form~\cite{wittenberg2021}.
Another experiment indicates that the video format makes news pieces perceived as more credible and more likely to be shared~\cite{sundar2021}.
Such an effect makes the misinformation video dangerous and it may lead to further negative impacts for various stakeholders.
For individuals, misinformation videos are more likely to mislead audiences to generate false memory and make misinformed decisions.
For online platforms, the wide spread of misinformation videos may lead to reputation crises, users' inactivity, and regulatory checks.
For watchdog bodies, misinformation videos may be weaponized to foment unrest and even undermine democratic institutions.
Therefore, actions are urgently required to reduce the risk brought by misinformation videos.

As a countermeasure, online platforms have made efforts to mitigate the spread of misinformation videos. 
For instance, TikTok introduces an enhanced in-app reporting feature to help curb the spread of COVID-19 misinformation~\cite{TiktokInterven}.
However, the current solution relies much on human efforts from expert teams or active users, which is labor-intensive and time-consuming.
It usually fails to perform real-time detection, and thus could not react rapidly when a new event emerges or previously debunked misinformation recurs~\cite{qi2022fakesv}. Moreover, this solution may introduce uncertain individual biases and errors~\cite{gupta2022combating}.
To tackle the above problems, developing techniques and systems for automatic misinformation video detection becomes a promising option.

Compared with text-based or text-image misinformation detection, video-based misinformation detection faces several unique challenges. 
First, the proliferation of heterogeneous information from diverse modalities brought more uncertainty and even noise to the final prediction.
Second, nonmalicious video-editing behaviors blur the distinction between forged and real videos.
Third, the recommendation-dominated content distribution of online video platforms reshapes the misinformation propagation from explicit behaviors like forwarding to implicit behaviors like re-uploading.
These challenges necessitate new technical solutions for detecting video-based misinformation and also highlight the importance of conducting a careful, specific investigation into this problem.

Despite many valuable surveys conducted on broad misinformation detection, limited attention is given to video-based misinformation.
Most of them regard the video as a kind of visual content as the image and discuss general multi-modal techniques for misinformation detection~\cite{cao2020exploring,abdali2022survey,alam2022survey}. The above-mentioned uniqueness of video-based misinformation is not sufficiently considered.
Other related surveys focus on a specific type of misinformation video, such as forged videos~\cite{agrawal2021survey,mirsky2021deepfakesurvey}, which provide detailed reviews but lack comprehensive analysis of the problem.
\yuyan{\citet{venkatagiri2023challenges} discussed the challenges and future research directions of studying misinformation on video-sharing platforms but did not provide an overview of specific detection methods.}
Considering the potential harms of misinformation videos, conducting a comprehensive survey on the detection problem is of urgent need.

To change the status quo and facilitate further exploration of this challenging problem by the research community, we present an overview of misinformation video detection in this survey.
Our main contributions are as follows:
\begin{itemize}[leftmargin=10pt]
    \item \textbf{Comprehensive characterization:} We present a comprehensive analysis of characteristics of misinformation videos from three levels including signals, semantics, and intents;
    \item \textbf{Systematic technical overview:} We provide a systematical overview of existing multi-modal detection techniques and methods for misinformation in the video form with a principled way to group utilized clues and integration mechanisms;
    \item \textbf{Concrete future directions:} We discuss several open issues in this area and provide concrete future directions for both research and real-world application scenarios.
\end{itemize}


\begin{figure*}[!t]  
    \centering    
    \includegraphics[width=0.95\linewidth]{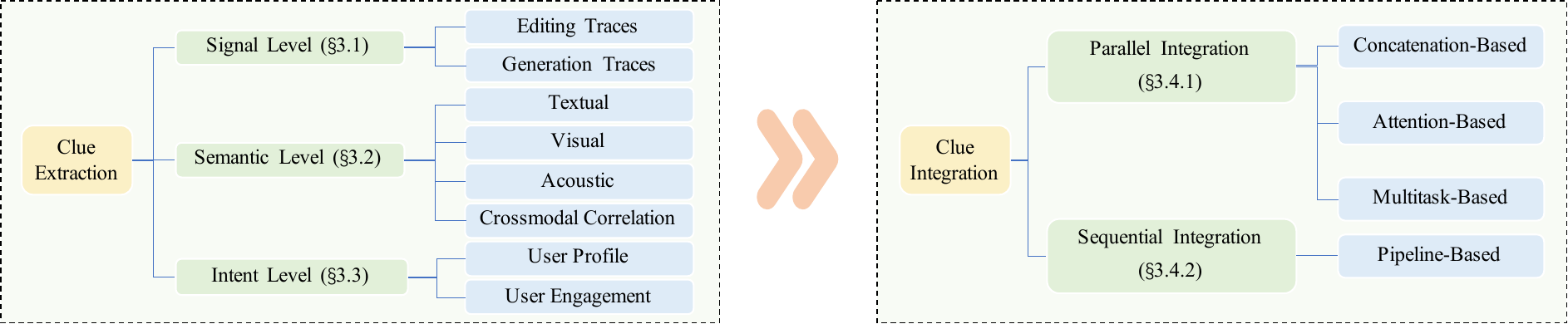}
    \caption{Overview of misinformation video detection techniques.}
    \label{fig:detection} 
\end{figure*}

\section{Misinformation Video Characterization}
\label{sec:char}
In this section, we characterize the misinformation video by giving the definition and analyzing it from three levels.
Following~\citet{zhou2020survey}, we define the misinformation video as:
\begin{definition}[Misinformation Video]  
    A video post that conveys false, inaccurate, or misleading information. 
\end{definition}
Note that a video post may include not only the video itself. On text-dominated social platforms like Facebook and Twitter, there could be a text paragraph attached; and on video-dominated platforms mentioned before, a title or a short text description is generally included.
Moreover, the concept of \textit{misinformation video} is closely related but different from \textit{fake video} because the former is characterized by the intention of spreading false or misleading information while the latter typically refers to the video that has been manipulated or produced by Photoshop-like tools and generative AI techniques. A fake video can be certainly used to create a misinformation video, but not all video faking is malicious (e.g., for movie production). Conversely, a video that is not manipulated but accompanied by false or misleading text will be considered a misinformation video. 
Therefore, detecting misinformation videos requires the appropriate combination of multi-perspective clues.

To find helpful detection clues, we characterize the misinformation video according to how it is produced. From the surface to the inside, the analysis is presented from three levels, including signal, semantic, and intent.
 
\subsection{Signal Level}
Misinformation videos often contain manipulated or generated video and audio content in which the forgery procedure often leads to traces in underlying digital signals.
Forgery methods that produce such traces can be classified into two groups: Editing and generation. Editing refers to visual alterations on existing data of video and audio modality. Typical editing actions include screen cropping, frame splicing, wave cutting, tempo changing, etc., which could be done using editing software~\cite{aneja2021mmsys}. Generation actions, by contrast, are done by neural networks which are trained to directly generate complete vivid videos (mostly with instructions from human actions~\cite{dancenow} or texts~\cite{make-a-video}). The generated videos may contain forged human faces or voices to mislead the audience.

\subsection{Semantic Level}
The falsehood is conveyed through incorrect semantic changes against the truth. For a misinformation video, such changes may occur in one specific modality or across multiple ones.
In the former case, manipulated elements (e.g., an exaggerated claim in the text description) are reflected by a single modality.
In the latter case, which is more common in multi-modal situations, misinformation might be conveyed by wrong semantic associations among non-forged contents of different modalities. For instance, a creator may upload a real video of an event that happened before but add a real text description of a newly emerging event.
\subsection{Intent Level}
\label{char-intent}

The creation of misinformation is often motivated by underlying intents, such as political influence, financial gain, and propaganda effects~\cite{MisinformationIntent}. To achieve the underlying intent, misinformation videos generally pursue wide and fast spread. This leads to unique patterns of expression, propagation, and user feedback, which are different from real ones. For example, \citet{qi2022fakesv} find that compared with real news videos, fake news videos have more significant emotional preferences, involve more user engagement, and are more likely to be questioned in comments.

\section{Misinformation Video Detection}
\label{sec:detect}
 The misinformation video detection problem is generally formulated as a binary classification task:
 Let $\mathcal{V}$ and $\mathcal{S}$ denote a \textsl{Video Post} and the attached \textsl{Social Context}, respectively. $\mathcal{V}$ consists of attributes such as title/description $t$, video $v$, and audio $a$. Social Context $S$ consists of two major components: User Profile $p$ and User Engagement $e$. User Profile $p$ includes a set of features to describe the uploader account, such as the location indicated by IP address, self-written and verified introduction, and follower count. User Engagement $e$ includes comments as well as statistics such as the number of views, likes, and stars. Fig.~\ref{fig:videopost} illustrates an example of $\mathcal{V}$ from TikTok.
The task of misinformation video detection is to predict whether the video post $\mathcal{V}$ contains misinformation given all the accessible features $\mathcal{E}=\{\mathcal{V},\mathcal{S}\}$, i.e., $\mathcal{F}:\mathcal{E} \mapsto \{0, 1\}$.
Existing studies usually take common metrics in classification tasks including accuracy, precision, recall, and F1 score for performance evaluation. 
Following the characterization in Sec.~\ref{sec:char}, we introduce a series of relevant works for detecting misinformation videos from clues at different levels to integration patterns. Fig.~\ref{fig:detection} gives an overview of this section.

\subsection{Signal Level} \label{signal_level}

Since misinformation videos are often created using forgery techniques, the detection of video forgery traces would provide a significant clue for detecting a misinformation video. Related works, which are always described as multimedia forensics, have received significant attention over the past decades. We introduce some commonly used techniques for the detection of editing traces and generation traces respectively. Given that there have been surveys on multimedia forensic techniques, here we will not go much into detail about the individual studies.
\subsubsection{Editing Traces}
As aforementioned, editing refers to alterations to the visual or audio content by multimedia editing software (mostly requiring manual operations). Existing detection methods for editing traces can be mainly categorized into two groups: active detection and passive detection. 

In active detection methods, digital watermarks or digital signatures are pre-embedded and extracted to detect the trace of secondary editing. For instance, \citet{tarhouni2022watermark} propose a blind and semi-fragile video watermarking scheme for detection. Combined watermarking (frames and audio watermarking) is used for detecting manipulation in both channels. 
In industry, the Coalition for Content Provenance and Authenticity develops a technical specification based on digital signature techniques for certifying the provenance and authenticity of media content. Compared to passive detection, active detection methods could generally provide quicker responses and more accurate judgments.

However, most of the videos are not pre-embedded with such additional information in practice. The passive detection methods highlight their advantages as they use the characteristics of the digital video itself to detect tampering traces. 
For video frames, tampering detection methods leverage inter-frame and intra-frame information. 
The former detects the abnormal change in frame sequences, such as insertion~\cite{zampoglou2019detecting}, deletion~\cite{abbasi2017malicious}, duplication~\cite{fadl2018authentication}, and shuffling~\cite{sitara2017comprehensive} of frames. The latter detects the alteration of objects that could be reflected in a single frame, such as region duplication~\cite{al2018detect_obj_move,antony2018video_copymove} and splicing~\cite{saddique2019spatial,johnston2020video}. Besides, artifacts acquired during the process of compression, noise artifacts left by the digital video camera, and inconsistencies (e.g., lighting, brightness, shadows) are also utilized as clues for detection~\cite{shelke2021comprehensive}.
For the audio content, statistical features inspired by observed properties are leveraged for forgery detection. Among them, Electric Network Frequency is widely used for forensics, thanks to its properties of random fluctuation around the nominal value and intra-grid consistency~\cite{reis2016esprit}.

\subsubsection{Generation Traces} \label{generation_traces}
Videos generated by neural networks (e.g., generative adversarial networks) are also known as ``deepfake videos''. Among them, deepfake videos containing vivid, generated human faces have been used for impersonating celebrities and brought negative impacts. The past few years have witnessed significant progress in detecting visual deepfakes. With reference to how deepfake videos are created~\cite{mirsky2021deepfakesurvey}, detection methods leverage clues generated at different steps including model fingerprints~\cite{yu2019attributing}, biological signals~\cite{matern2019exploiting}, and temporal consistency~\cite{yang2019exposing,haliassos2021lips}. 
With recent advances in Text-To-Speech and Voice Conversion algorithms, generating fake audio that is indistinguishable for humans become easier and attracted increasing attention. ASVspoof challenges~\cite{yamagishi2022lessons} were organized for accelerating progress in deepfake audio detection. Most existing works for fake audio detection adopt hand-crafted acoustic features like Mel-frequency cepstral coefficients (MFCC), linear frequency cepstral coefficient (LFCC), and constant-Q cepstral coefficients (CQCC) and apply classifiers such as Gaussian mixture model and light convolutional neural network to make predictions. Recent work also attempts to leverage pre-trained wav2vec-style models for speech representation extraction and build an end-to-end framework~\cite{lv2022fake}.

Joint audio-visual deepfake detection is to handle the case that the visual or/and auditory modalities have been manipulated. Detection methods focus on learning intrinsic (a)synchronization between video and audio frames in real and fake videos~\cite{zhou2021jointdf,mittal2020emotions}. Moreover, different modalities can also supplement each other for final judgments~\cite{khalid2021evaluation}.

Considering that the action of video compression is widely used in the default upload setting of online video platforms, handling compressed deepfake videos becomes an important issue. The compression operation erases certain generation traces and thus makes videos more undetectable. Moreover, the quantization and inverse quantization operations during compression and decompression bring additional quantization noise and distortion. To tackle these problems, \citet{hu2021detecting} propose a two-stream method by analyzing the frame-level and temporality-level of compressed deepfake videos for detection.
\subsection{Semantic Level} \label{semantic_level}
Though signal-level clues could provide strong evidence, they are not decisive because of the wide use of portable editing tools. Even if editing or generation traces are detected, it does not necessarily mean that the video conveys misinformation. The existence of semantically unchanged edited videos has blurred the boundary between fake and real videos. For instance, a video that has been edited for the sake of brevity or clarity may still be truthful and informative. Conversely, a video that is technically untampered can be employed in a deceptive manner. Different from clues at the signal level, semantic-level clues offer a new perspective for identifying misinformative content.
Thus, many recent works focus on leveraging multi-modal semantic clues from not only the video content but also descriptive textual information. In this section, we discuss the features exploited by semantic-based methods from a multi-modal perspective.

\textbf{Textual Feature.} The video content is always served with descriptive textual information such as video description, and title. Apart from these directly accessible texts, subtitles and transcriptions extracted from the video also present useful information. Early works most extract statistical features from these texts for classification. \citet{papadopoulou2017} first exploit linguistic features of the title, which contain basic attributes like text length, as well as designed indicators like whether a title contains specified characters (e.g., the question/exclamation mark and 1st/2nd/3rd person pronoun), the number of special words (e.g., slang words and sentiment-indicative words) and the readability score. Other works also consider the existence of specific expressions like clickbait phrases and violent words for detection\cite{palod2019vavd,li2022cnn}. Corpus-aware features, such as n-grams, TF-IDF, lexical richness, and LIWC lexicon overlapping, are leveraged by \citet{hou2019icmi} and \citet{aclworkshop2020}. In addition to hand-crafted features, continuous representation generated using deep learning has been increasingly adopted. \citet{jagtap2021misinformation} employs GloVe~\cite{glove} and Word2Vec~\cite{word2vec} to generate subtitle embeddings. \citet{shang2021tiktec} and \citet{choi2021cikm} train bidirectional recurrent neural networks as text encoders to encode the semantics of textual information including the title, description, and transcription. Recent advances in pre-trained language models (e.g., BERT~\cite{bert}) also drive the latest multi-modal detection models to obtain contextualized representation\yuyan{~\cite{choi2021cikm,wang2022misinformation,mccrae2022multi,qi2022fakesv,christodoulou2023identifying}}. 
Moreover, factual elements like event triggers and event augments are utilized to provide explicit guidance to learn the internal event semantics in~\cite{liu2023covid}.

\textbf{Visual Feature.} The visual content is usually represented at the frame level or clip level. The former presents static visual features while the latter presents additional temporal features. \citet{shang2021tiktec} extract frames through uniform sampling and input the resized sampled frames into the advanced object detection network Fast R-CNN for visual features of object regions. Corresponding caption representation is used to guide the integration of object regions to help generate the frame visual representation. \citet{choi2021cikm} extract frames according to their similarity to the thumbnail. Pre-trained VGG-19 is utilized to extract visual features from the video frames. \citet{mccrae2022multi} break video into 32-frames-long clips with each clip beginning at a keyframe. The keyframes are detected through the FFmpeg scene detection filter. For each clip, features related to human faces, objects, and activities are extracted through pre-trained FaceNet, ResNet50, and S3D networks, respectively. \citet{wang2022misinformation} break the video into clips with fixed duration directly and uses S3D to extract visual features likewise. \citet{qi2022fakesv} represent visual content both at the frame level and clip level. The pre-trained VGG19 model and pre-trained C3D model are used to extract frame features and clip features respectively. \citet{liu2023covid} encode frames into features using the pre-trained vision transformer ViT~\citep{vit} used in CLIP~\citep{clip}.

\textbf{Acoustic Feature.}  As a unique modality compared to text-image misinformation, the audio modality including speech, environment sound, and background music~\cite{qi2022fakesv}, plays an essential role in expressing information in videos. As for detection, in addition to the transcription mentioned above, current works search for useful clues from acoustic characteristics. \citet{hou2019icmi} firstly import emotional acoustic features to the detection model, where predefined feature sets widely used for emotion recognition of raw speech are exploited.
\citet{shang2021tiktec} design an acoustic-aware speech encoder by introducing MFCC features. \citet{qi2022fakesv} use the pre-trained VGGish to extract the audio features for classification. 

\textbf{Cross-modal Correlation.} Mismatches between modalities, such as video-text and video-audio, are often caused by video repurposing for the misleading aim, which would lead to important changes in cross-modal correlation.
\citet{liu2023covid} apply the cross-modal transformer to learn the consistent relationship between video and speech, video and text, and speech and text, respectively. The pairwise consistency scores are then aggregated for final judgments.
\citet{mccrae2022multi} leverage an ensemble method based on textual analysis of the caption, automatic audio transcription, semantic video analysis, object detection, named entity consistency, and face verification for mismatch identification. \citet{wang2022misinformation} propose two methods based on contrastive learning and masked language modeling for joint representation learning to identify semantic inconsistencies. In~\cite{choi2021cikm}, topic distribution differences between modalities are utilized to robustify the detection model.

The increasing diversity of semantic features has inspired the evolution of misinformation video detection from mining unimodal patterns to modeling multi-modal interactions. However, despite the progress, existing works still have some limitations. For instance, most of them extract features from the perspective of identifying misinformative patterns, which may result in overlooking factual information and failing to utilize clues that require external knowledge for reasoning and judgment. \yuyan{Despite the pioneering efforts of \cite{qi2023two} in incorporating relevant factual information by rectifying the wrong predictions of previously detected news videos with reliable debunking videos, there remains an under-exploration of multi-modal verification.} Moreover, logical constraints, such as inter-modality consistency and entailment, have not been sufficiently utilized, unlike in text-image modeling~\cite{qi-mm2021,sun-etal-2021-inconsistency-matters}. These limitations suggest that further improvements are required to enhance the accuracy and robustness of misinformation video detection methods.

\subsection{Intent Level}
Misinformation videos are often created and shared with deliberate intents, e.g., for financial and political gains~\cite{shu2017survey} or self-expression~\cite{user-self-description}. Compared with clues at the signal and semantic levels, clues at the intent level are usually more robust to elaborately produced misinformation videos that avoid being detected, because the underlying essential intent and its effects on the behaviors of social media users are less likely to be changed or manipulated at scale. 
Starting from the motivations of those who create and spread misinformation, some effective features for detecting misinformation videos are leveraged by researchers.

Social contexts refer to user social engagements and profiles when information spreads on platforms with social media characteristics.
As mentioned in Sec.~\ref{char-intent}, unique social contexts might reflect the spreading intent of misinformation creators and thus provide useful features~\cite{qi2022fakesv}.
Current works mostly make use of user comments and statistics on user engagement. The comments are usually exploited by extracting hand-crafted features~\cite{papadopoulou2017} or generating general representation vectors through deep models~\cite{palod2019vavd,choi2021cikm,qi2022fakesv}. Some works go deeper in mining comments. For example, \citet{aclworkshop2020} learn a feature of comment conspiracy and \citet{choi2022prl} give an eye to the domain knowledge. User engagement statistics such as the number of likes, comments, and views are generally directly concatenated with other features before being put into the classifier~\cite{hou2019icmi}. Some work also uses statistical numbers as importance weights to help generate embedding.
\citet{choi2022prl} generate video comment embeddings by calculating the weighted sum of embeddings of each comment using their numbers of likes.
The publisher profile provides auxiliary information about source credibility in post-level detection. \citet{papadopoulou2017} and \citet{li2022cnn} leverage a series of features, such as the number of views of the publisher channel, number of published videos, and follower-following ratio. \citet{qi2022fakesv} also point out that user profiling features like geolocation information and whether the account is verified or not can be useful to the detection, and exploit the textual publisher description in their model.

Note that social context is not the only choice to capture intent-level clues. In text-based and text-image studies, researchers have started preliminary trials to mine the intents by directly analyzing the content itself~\cite{intent1,intent2}.

\begin{table*}
\centering
\small
\setlength{\tabcolsep}{3.2pt}
\setlength{\extrarowheight}{0pt}
\addtolength{\extrarowheight}{\aboverulesep}
\addtolength{\extrarowheight}{\belowrulesep}
\setlength{\aboverulesep}{0pt}
\setlength{\belowrulesep}{0pt}
\caption{Summary of misinformation video detection methods. E/G: Editing/Generation Traces. T: Textual. V: Visual. A: Acoustic. CMC: Cross-modal Correlation. UE: User Engagements. UP: User Profile. C: Concatenation-Based. ATT: Attention-Based. MT: Multitask-Based. PL: Pipeline-Based. Methods for fake video detection are not included due to the space limit.}
\label{tab:detection}
\resizebox{0.85\textwidth}{!}{
\begin{tabular}{lccrcccccccccccc} 
\toprule
\multicolumn{2}{c}{\multirow{2}{*}[0em]{\textbf{Method}}} & \multicolumn{3}{c}{\textbf{Dataset}} &  \multicolumn{7}{c}{{\cellcolor[rgb]{0.937,0.937,0.937}} \textbf{Clue Extraction}} & \multicolumn{4}{c}{{\cellcolor[rgb]{0.855,0.855,0.859}} \textbf{Clue Integration}}\\ 
&  & \multirow{1}{*}[0em]{\textbf{Source}} & \multicolumn{1}{c}{\multirow{1}{*}[0em]{\textbf{Amount}}} & \multirow{1}{*}[0em]{\textbf{Released?}} & {\cellcolor[rgb]{0.937,0.937,0.937}}\textbf{E/G} & {\cellcolor[rgb]{0.937,0.937,0.937}}\textbf{T}& {\cellcolor[rgb]{0.937,0.937,0.937}}\textbf{V} & {\cellcolor[rgb]{0.937,0.937,0.937}}\textbf{A}& {\cellcolor[rgb]{0.937,0.937,0.937}} \textbf{CMC}& {\cellcolor[rgb]{0.937,0.937,0.937}}\textbf{UE} & {\cellcolor[rgb]{0.937,0.937,0.937}}\textbf{UP} & {\cellcolor[rgb]{0.855,0.855,0.859}}\textbf{C} & {\cellcolor[rgb]{0.855,0.855,0.859}}\textbf{ATT} & {\cellcolor[rgb]{0.855,0.855,0.859}}\textbf{MT} & {\cellcolor[rgb]{0.855,0.855,0.859}}\textbf{PL}  \\ 
\midrule
\citet{liu2023covid}        & 2023   & Twitter    & 10,000   & N   & {\cellcolor[rgb]{0.937,0.937,0.937}}      & {\cellcolor[rgb]{0.937,0.937,0.937}}$\bullet$  & {\cellcolor[rgb]{0.937,0.937,0.937}}$\bullet$  & {\cellcolor[rgb]{0.937,0.937,0.937}}    & {\cellcolor[rgb]{0.937,0.937,0.937}}$\bullet$  & {\cellcolor[rgb]{0.937,0.937,0.937}}    & {\cellcolor[rgb]{0.937,0.937,0.937}}    & {\cellcolor[rgb]{0.855,0.855,0.859}}    & {\cellcolor[rgb]{0.855,0.855,0.859}}$\bullet$   & {\cellcolor[rgb]{0.855,0.855,0.859}}    & {\cellcolor[rgb]{0.855,0.855,0.859}}    \\
\citet{qi2022fakesv,qi2023two}          & 2023   & Douyin, Kuaishou    & 5,538    & Y   & {\cellcolor[rgb]{0.937,0.937,0.937}}      & {\cellcolor[rgb]{0.937,0.937,0.937}}$\bullet$  & {\cellcolor[rgb]{0.937,0.937,0.937}}$\bullet$  & {\cellcolor[rgb]{0.937,0.937,0.937}}$\bullet$  & {\cellcolor[rgb]{0.937,0.937,0.937}}$\bullet$  & {\cellcolor[rgb]{0.937,0.937,0.937}}$\bullet$  & {\cellcolor[rgb]{0.937,0.937,0.937}}$\bullet$   & {\cellcolor[rgb]{0.855,0.855,0.859}}    & {\cellcolor[rgb]{0.855,0.855,0.859}}$\bullet$   & {\cellcolor[rgb]{0.855,0.855,0.859}}    & {\cellcolor[rgb]{0.855,0.855,0.859}}    \\
\citet{ganti2022}  & 2022   & Not Specified     & -    & N   & {\cellcolor[rgb]{0.937,0.937,0.937}}$\bullet$    & {\cellcolor[rgb]{0.937,0.937,0.937}}$\bullet$  & {\cellcolor[rgb]{0.937,0.937,0.937}}  & {\cellcolor[rgb]{0.937,0.937,0.937}}    & {\cellcolor[rgb]{0.937,0.937,0.937}}       & {\cellcolor[rgb]{0.937,0.937,0.937}}    & {\cellcolor[rgb]{0.937,0.937,0.937}}    & {\cellcolor[rgb]{0.855,0.855,0.859}}    & {\cellcolor[rgb]{0.855,0.855,0.859}}     & {\cellcolor[rgb]{0.855,0.855,0.859}}    & {\cellcolor[rgb]{0.855,0.855,0.859}}$\bullet$   \\
\citet{mccrae2022multi}   & 2022   & Facebook, YouTube    & 4,651    & N   & {\cellcolor[rgb]{0.937,0.937,0.937}}      & {\cellcolor[rgb]{0.937,0.937,0.937}}$\bullet$  & {\cellcolor[rgb]{0.937,0.937,0.937}}$\bullet$  & {\cellcolor[rgb]{0.937,0.937,0.937}}    & {\cellcolor[rgb]{0.937,0.937,0.937}}$\bullet$  & {\cellcolor[rgb]{0.937,0.937,0.937}}$\bullet$  & {\cellcolor[rgb]{0.937,0.937,0.937}}    & {\cellcolor[rgb]{0.855,0.855,0.859}}$\bullet$   & {\cellcolor[rgb]{0.855,0.855,0.859}}     & {\cellcolor[rgb]{0.855,0.855,0.859}}    & {\cellcolor[rgb]{0.855,0.855,0.859}}    \\
\citet{wang2022msmd}  & 2022   & Twitter    & 943,667  & N   & {\cellcolor[rgb]{0.937,0.937,0.937}}      & {\cellcolor[rgb]{0.937,0.937,0.937}}$\bullet$  & {\cellcolor[rgb]{0.937,0.937,0.937}}$\bullet$  & {\cellcolor[rgb]{0.937,0.937,0.937}}$\bullet$  & {\cellcolor[rgb]{0.937,0.937,0.937}}$\bullet$  & {\cellcolor[rgb]{0.937,0.937,0.937}}    & {\cellcolor[rgb]{0.937,0.937,0.937}}    & {\cellcolor[rgb]{0.855,0.855,0.859}}    & {\cellcolor[rgb]{0.855,0.855,0.859}}$\bullet$   & {\cellcolor[rgb]{0.855,0.855,0.859}}$\bullet$   & {\cellcolor[rgb]{0.855,0.855,0.859}}    \\
\citet{wang2022misinformation}         & 2022   & Twitter    & 160,000  & N   & {\cellcolor[rgb]{0.937,0.937,0.937}}      & {\cellcolor[rgb]{0.937,0.937,0.937}}$\bullet$  & {\cellcolor[rgb]{0.937,0.937,0.937}}$\bullet$  & {\cellcolor[rgb]{0.937,0.937,0.937}}    & {\cellcolor[rgb]{0.937,0.937,0.937}}$\bullet$  & {\cellcolor[rgb]{0.937,0.937,0.937}}    & {\cellcolor[rgb]{0.937,0.937,0.937}}    & {\cellcolor[rgb]{0.855,0.855,0.859}}    & {\cellcolor[rgb]{0.855,0.855,0.859}}$\bullet$   & {\cellcolor[rgb]{0.855,0.855,0.859}}$\bullet$   & {\cellcolor[rgb]{0.855,0.855,0.859}}    \\
\citet{li2022cnn}  & 2022   & Bilibili   & 700  & N   & {\cellcolor[rgb]{0.937,0.937,0.937}}      & {\cellcolor[rgb]{0.937,0.937,0.937}}$\bullet$  & {\cellcolor[rgb]{0.937,0.937,0.937}}  & {\cellcolor[rgb]{0.937,0.937,0.937}}    & {\cellcolor[rgb]{0.937,0.937,0.937}}       & {\cellcolor[rgb]{0.937,0.937,0.937}}$\bullet$  & {\cellcolor[rgb]{0.937,0.937,0.937}}$\bullet$   & {\cellcolor[rgb]{0.855,0.855,0.859}}$\bullet$   & {\cellcolor[rgb]{0.855,0.855,0.859}}     & {\cellcolor[rgb]{0.855,0.855,0.859}}    & {\cellcolor[rgb]{0.855,0.855,0.859}}    \\
\citet{choi2022prl}  & 2022   & YouTube    & 2,912    & N   & {\cellcolor[rgb]{0.937,0.937,0.937}}      & {\cellcolor[rgb]{0.937,0.937,0.937}}$\bullet$  & {\cellcolor[rgb]{0.937,0.937,0.937}}$\bullet$  & {\cellcolor[rgb]{0.937,0.937,0.937}}    & {\cellcolor[rgb]{0.937,0.937,0.937}}       & {\cellcolor[rgb]{0.937,0.937,0.937}}$\bullet$  & {\cellcolor[rgb]{0.937,0.937,0.937}}    & {\cellcolor[rgb]{0.855,0.855,0.859}}$\bullet$   & {\cellcolor[rgb]{0.855,0.855,0.859}}     & {\cellcolor[rgb]{0.855,0.855,0.859}}    & {\cellcolor[rgb]{0.855,0.855,0.859}}    \\
\citet{choi2021cikm}           & 2021   & YouTube    & 4,622    & N   & {\cellcolor[rgb]{0.937,0.937,0.937}}      & {\cellcolor[rgb]{0.937,0.937,0.937}}$\bullet$  & {\cellcolor[rgb]{0.937,0.937,0.937}}$\bullet$  & {\cellcolor[rgb]{0.937,0.937,0.937}}    & {\cellcolor[rgb]{0.937,0.937,0.937}}$\bullet$  & {\cellcolor[rgb]{0.937,0.937,0.937}}$\bullet$  & {\cellcolor[rgb]{0.937,0.937,0.937}}    & {\cellcolor[rgb]{0.855,0.855,0.859}}$\bullet$   & {\cellcolor[rgb]{0.855,0.855,0.859}}     & {\cellcolor[rgb]{0.855,0.855,0.859}}$\bullet$   & {\cellcolor[rgb]{0.855,0.855,0.859}}    \\
\citet{shang2021tiktec}        & 2021   & YouTube    & 891  & N   & {\cellcolor[rgb]{0.937,0.937,0.937}}      & {\cellcolor[rgb]{0.937,0.937,0.937}}$\bullet$  & {\cellcolor[rgb]{0.937,0.937,0.937}}$\bullet$  & {\cellcolor[rgb]{0.937,0.937,0.937}}$\bullet$  & {\cellcolor[rgb]{0.937,0.937,0.937}}       & {\cellcolor[rgb]{0.937,0.937,0.937}}    & {\cellcolor[rgb]{0.937,0.937,0.937}}    & {\cellcolor[rgb]{0.855,0.855,0.859}}    & {\cellcolor[rgb]{0.855,0.855,0.859}}$\bullet$   & {\cellcolor[rgb]{0.855,0.855,0.859}}    & {\cellcolor[rgb]{0.855,0.855,0.859}}    \\
\citet{jagtap2021misinformation}       & 2021   & YouTube    & 2125     & Y   & {\cellcolor[rgb]{0.937,0.937,0.937}}      & {\cellcolor[rgb]{0.937,0.937,0.937}}$\bullet$  & {\cellcolor[rgb]{0.937,0.937,0.937}}  & {\cellcolor[rgb]{0.937,0.937,0.937}}    & {\cellcolor[rgb]{0.937,0.937,0.937}}       & {\cellcolor[rgb]{0.937,0.937,0.937}}    & {\cellcolor[rgb]{0.937,0.937,0.937}}    & \multicolumn{1}{c}{{\cellcolor[rgb]{0.855,0.859,0.859}}}           & \multicolumn{1}{c}{{\cellcolor[rgb]{0.855,0.859,0.859}}}        & \multicolumn{1}{c}{{\cellcolor[rgb]{0.855,0.859,0.859}}}           & \multicolumn{1}{c}{{\cellcolor[rgb]{0.855,0.859,0.859}}}           \\
\citet{aclworkshop2020}        & 2020   & YouTube    & 180  & N   & {\cellcolor[rgb]{0.937,0.937,0.937}}      & {\cellcolor[rgb]{0.937,0.937,0.937}}$\bullet$  & {\cellcolor[rgb]{0.937,0.937,0.937}}  & {\cellcolor[rgb]{0.937,0.937,0.937}}    & {\cellcolor[rgb]{0.937,0.937,0.937}}       & {\cellcolor[rgb]{0.937,0.937,0.937}}$\bullet$  & {\cellcolor[rgb]{0.937,0.937,0.937}}    & {\cellcolor[rgb]{0.855,0.855,0.859}}$\bullet$   & {\cellcolor[rgb]{0.855,0.855,0.859}}     & {\cellcolor[rgb]{0.855,0.855,0.859}}    & {\cellcolor[rgb]{0.855,0.855,0.859}}    \\
\citet{hou2019icmi}  & 2019   & YouTube    & 250  & N   & {\cellcolor[rgb]{0.937,0.937,0.937}}      & {\cellcolor[rgb]{0.937,0.937,0.937}}$\bullet$  & {\cellcolor[rgb]{0.937,0.937,0.937}}  & {\cellcolor[rgb]{0.937,0.937,0.937}}$\bullet$  & {\cellcolor[rgb]{0.937,0.937,0.937}}       & {\cellcolor[rgb]{0.937,0.937,0.937}}$\bullet$  & {\cellcolor[rgb]{0.937,0.937,0.937}}    & {\cellcolor[rgb]{0.855,0.855,0.859}}$\bullet$   & {\cellcolor[rgb]{0.855,0.855,0.859}}     & {\cellcolor[rgb]{0.855,0.855,0.859}}    & {\cellcolor[rgb]{0.855,0.855,0.859}}    \\
\citet{palod2019vavd}          & 2019   & YouTube    & 546  & Y   & {\cellcolor[rgb]{0.937,0.937,0.937}}      & {\cellcolor[rgb]{0.937,0.937,0.937}}$\bullet$  & {\cellcolor[rgb]{0.937,0.937,0.937}}  & {\cellcolor[rgb]{0.937,0.937,0.937}}    & {\cellcolor[rgb]{0.937,0.937,0.937}}       & {\cellcolor[rgb]{0.937,0.937,0.937}}$\bullet$  & {\cellcolor[rgb]{0.937,0.937,0.937}}    & {\cellcolor[rgb]{0.855,0.855,0.859}}$\bullet$   & {\cellcolor[rgb]{0.855,0.855,0.859}}     & {\cellcolor[rgb]{0.855,0.855,0.859}}    & {\cellcolor[rgb]{0.855,0.855,0.859}}    \\
\citet{papadopoulou2017}           & 2017   & YouTube, Twitter, Facebook  & 5,006    & Y   & {\cellcolor[rgb]{0.937,0.937,0.937}}      & {\cellcolor[rgb]{0.937,0.937,0.937}}$\bullet$  & {\cellcolor[rgb]{0.937,0.937,0.937}}  & {\cellcolor[rgb]{0.937,0.937,0.937}}    & {\cellcolor[rgb]{0.937,0.937,0.937}}       & {\cellcolor[rgb]{0.937,0.937,0.937}}$\bullet$  & {\cellcolor[rgb]{0.937,0.937,0.937}}$\bullet$   & {\cellcolor[rgb]{0.855,0.855,0.859}}$\bullet$   & {\cellcolor[rgb]{0.855,0.855,0.859}}     & {\cellcolor[rgb]{0.855,0.855,0.859}}    & {\cellcolor[rgb]{0.855,0.855,0.859}}    \\
\bottomrule
\end{tabular}}
\end{table*}

\subsection{Clue Integration}
In previous subsections, we categorized the various features utilized in existing works for detecting misinformation videos at the signal, semantic, and intent levels. In practice, existing methods generally combine multiple features from different modalities to make a judgment. In this section, we group the methods for integrating clues into two types (i.e., parallel and sequential integration) and introduce the current advances.
\subsubsection{Parallel Integration}
In parallel integration, all clues from different modalities contribute to the final decision-making process, although their participation may not be equal. The integration process can be performed at both the feature level (where features from different modalities are fused before being input into the classifier) and the decision level (where predictions are produced independently by different branches and combined using strategies like voting for final decisions). In existing works, feature fusion is used more frequently than decision integration. Here we discuss three types of feature fusion techniques used in existing works.

\textbf{Concatenation-Based}: The majority of the existing works on multi-modal misinformation detection embed each modality into a representation vector and then concatenate them as a multi-modal representation. The generated representation can be utilized for classification tasks directly or input into a deep network (e.g., the convolutional neural network) for deeper fusion and classification~\cite{li2022cnn}. Linear combination is another simple but effective way to combine feature vectors of different modalities~\cite{choi2021cikm}.
The fusion process that combines features from different modalities can be done at the video level or at the frame/clip level~\cite{mccrae2022multi}.

\textbf{Attention-Based}: The attention mechanism is a more effective approach for utilizing embeddings of different modalities, as it jointly exploits the multi-modal feature by focusing on specific parts and allows dynamic fusion for sequential data. \citet{shang2021tiktec} use a co-attention module that simultaneously learns the pairwise relation between each pair of a video frame and spoken word to fuse the visual and speech information. \citet{wang2022misinformation} model the joint distribution of video and text by using a variant of masked language modeling. A transformer is trained to predict each text token given its text context and the video. \citet{qi2022fakesv} and \citet{wang2022msmd} utilize a cross-modal transformer to model the mutual interaction between different modalities.

\textbf{Multitask-Based}: Another utilized fusion architecture is based on multitask learning. Under this architecture, auxiliary networks are applied to learn individual or multi-modal representations, spaces, or parameters better and improve the classification performance~\cite{abdali2022survey}. For example, \citet{choi2021cikm} use a topic-adversarial classification to guide the model to learn topic-agnostic features for good generalization. \citet{wang2022misinformation} use contrastive learning to build the joint representation space of video and text.

\yuyan{In practice, no single method demonstrates overwhelming superiority. Concatenation-based fusion preserves all available information but it treats all features equally without explicitly considering their relative importance. Attention-based fusion enables the model to focus on more informative parts, but it necessitates more complex mechanism design and computing resources. Multitask-based fusion enhances the model's generalization capabilities but requires a larger amount of labeled data as well as more computing resources. The choice of fusion method should depend on the specific requirements of the task, available resources, and the trade-off between simplicity, computational demands, and performance improvement.}

\subsubsection{Sequential Integration}
In sequential integration, clues from different modalities are combined in a step-wise manner with each modality contributing incrementally to the final decision. This way of integration can promote overall efficiency and effectiveness, especially when some modalities provide redundant or overlapping information. We exemplified how sequential integration can be utilized in detecting misinformation videos with the two-pronged method proposed by~\citet{ganti2022}. This method operates by finding the original video corresponding to the given suspicious video and measuring the similarity between them. Initially, the method uses the reverse image search to retrieve the original video and calculates the similarity between frames of two videos to judge if the given video is a face-swapped deepfake video. If so, the video will be judged as a misinformative one; otherwise, the method analyzes the semantic similarity of video captions to detect the shifts in the meaning and intent behind the two videos. High semantic similarity leads to a real video judgment. To avoid the impact of sentiment changes on semantic similarity, the method compares video captions' sentiments for videos obtaining low semantic similarities. If the sentiments are similar, the method confirms that the semantic difference is not influenced by sentiments and judges the given video as a misinformative one. 

\autoref{tab:detection} summarizes misinformation video detection methods and specifies the data sources, exploited clues, and clue integration techniques for each method. Most of the methods exploit multiple clues and use concatenation and attention for clue integration. Including more modalities and better modeling inter-modality relationships are promising for improving misinformation video detection.

\section{Resources}
\label{sec:resource}
Resources are often the limiting factors for conducting research on this task. Here we introduce relevant datasets and tools and highlight their features and application contexts.

\subsection{Datasets}
\label{sec:dataset}
Due to the difficulty of video crawling (mostly based on carefully selected keywords) and human annotation, many existing datasets are small-scale and topic-specific.
\yuyan{We have compiled information about these datasets in Appendix~\ref{sec:appendix_dataset}, including the data source, topic domain, size, accessibility, and state-of-the-art (SOTA) methods (along with their performances) on these datasets.}
Unfortunately, most of the datasets are not released. Here we detail four large and publicly available datasets:

{
\begin{itemize}[leftmargin=10pt]
    \item \textbf{FVC\footnote{\url{https://mklab.iti.gr/results/fake-video-corpus/}}.} The initial FVC comprises videos from a variety of event categories (e.g., politics and sports), and contains 200 fake and 180 real videos. Using the initially collected videos as seeds and searching on three platforms (YouTube, Facebook, and Twitter), researchers extend FVC to a multi-lingual dataset containing 3,957 fake and 2,458 real videos, with textual news content and user comments attached.
    \item \textbf{YouTubeAudit\footnote{\url{https://social-comp.github.io/YouTubeAudit-data/}}.} This dataset contains 2,943 videos that were published in 2020 on YouTube and cover five popular misinformative topics~\cite{hussein2020audit}. Each sample is labeled as promoting, debunking, and neutral to misinformation. It also provides social contexts like metadata (e.g., video URL, title, duration), statistics of user engagements, and user profiles (e.g., gender and age).
    \item \textbf{FakeSV\footnote{\url{https://github.com/ICTMCG/FakeSV}}.} This dataset contains 5,538 Chinese short videos (1,827 fake, 1,827 real, and 1,884 debunking videos) crawled from the short video platforms Douyin and Kuaishou. It covers 738 news events from 2019 to 2022. It also provides social contexts including user responses and publisher profiles.
    \item \textbf{COVID-VTS\footnote{\url{https://github.com/FuxiaoLiu/Twitter-Video-dataset}}.} This dataset contains 10k COVID-related video-text pairs and is specifically for research on inter-modality consistency. Half of the samples are pristine and others are generated by partially modifying or replacing factual information in pristine samples.
\end{itemize}
}

\subsection{Tools}
Tools for specific utilities are valuable for verifying suspicious videos because they provide important auxiliary information that could be hardly learned by a data-driven model. We list three representative publicly available tools for different application contexts:

\begin{itemize}[leftmargin=10pt]
    \item \textbf{DeepFake Detector}: An AI service to judge whether a given video contains deepfake manipulated faces, developed within WeVerify Project\footnote{\url{https://weverify.eu/tools/deepfake-detector/}}. It uses the URL of a suspicious image or video as the input and returns the deepfake probability score. 
    It could help detect generation traces discussed in Sec.~\ref{generation_traces}.
    \item \textbf{Reverse Image Search}: Web search services with an image as a query. 
    By submitting extracted keyframes, we could check if there exist similar videos elsewhere before. This could help detect the manipulation by comparison~\cite{ganti2022} and identify the original source, as discussed in Secs.~\ref{signal_level} and~\ref{semantic_level}. Many general search engines (e.g., Google and Baidu) provide such services. Task-specific tools include TinEye, ImageRaider, and Duplichecker\footnote{\url{https://tineye.com/}, \url{https://infringement.report/api/raider-reverse-image-search/}, \& \url{https://www.duplichecker.com/}}.
    \item \textbf{Video Verification Plugin}: A Google Chrome extension to verify videos provided by the InVID European Project\footnote{\url{https://www.invid-project.eu/tools-and-services/invid-verification-plugin/}}. It provides a toolbox to obtain contextual information from YouTube or Facebook, extract keyframes for reverse search, show metadata, and perform forensic analysis, which is useful for the detection at the signal and semantic levels (Secs.~\ref{signal_level} and~\ref{semantic_level}).
\end{itemize}

\section{Related Areas}
\label{sec:relatedarea}

\yuyan{In this section, we discuss three areas related to misinformation video detection and clarify the relationships between misinformation video detection and these areas (Table~\ref{tab:compare-related-area}).}

\textbf{Deception Detection}: It aims at identifying the existence of deceptive behaviors, which is crucial for personal and public safety.
In earlier research, both verbal and nonverbal cues play important roles in deception detection~\cite{alaskar2022deceptionsurvey}. Verbal cues mainly refer to the linguistic characteristics of the statement while non-verbal cues include neurological, visual, and vocal indicators. 

\textbf{Harmful Content Detection}:
Harmful content generally renders as doxing, identity attack, identity misrepresentation, insult, sexual aggression, and the threat of violence~\cite{banko2020unified}. Detecting video-based harmful content often relies on capturing indicative features in multiple modalities, such as linguistic features of audio transcription, video sentiment, and flagged harmful objects~\cite{alam2022survey}. 

\textbf{Clickbait Detection}:
Clickbait is a term commonly used to describe eye-catching and teaser headlines (thumbnails) in online media~\cite{shu2017survey}. Clickbait video detection is to determine whether a video is faithfully representing the event it refers to. 
Content-based methods focus on analyzing the semantic gaps between the initially presented information (e.g., title and video thumbnail) and that expressed by the whole video, while others exploited creator profiles and audience feedback~\cite{varshney2021unified}. 

\begin{table}
\centering
\caption{Misinformation video detection vs. related areas.}
\label{tab:compare-related-area}
\resizebox{0.48\textwidth}{!}{
\begin{tabular}{m{0.18\textwidth}<{\centering}m{0.31\textwidth}} 
\toprule
Misinformation Detection vs. Deception Detection    &  Deceptive techniques could be used to convey fabricated information, but non-intentionally created misinformation may not correspond to deceptive behaviors.  \\ 
\midrule
Misinformation Detection vs. Harmful Content Detection & Harmful content is more likely to cause mental harm to specific populations or figures, while misinformation is meant to mislead readers.\\ 
\midrule
Misinformation Detection vs. Clickbait Detection & Clickbait could be seen as a manifestation of misinformation materials.\\
\bottomrule
\end{tabular}
}
\end{table}

\section{Open Issues and Future Directions}
\label{sec:future}
Though existing works have demonstrated significant advances in detecting misinformation videos, there are still many issues that barrier their application to real-world systems. Here we present two open issues and two future directions along with concrete tasks to advance the landscape of practical detection systems.

\subsection{Open Issues}
\subsubsection{Transferability}

Transferability reflects how well a detection system tackles data distribution shift which is common and inevitable in real-world applications. Despite being a hot research topic, this issue remains largely underexplored in misinformation video detection, which is a crucial barrier for detection methods to be put into practice.
Here we provide three transfer-related subproblems in different aspects:

\textit{1) Multi-platform detection.} The differences in contents and user groups among platforms shape different social contexts and provide extra clues, which has been found helpful in detection~\cite{qi2022fakesv,micallef2022cross}. This is important for this area given that cross-platform sharing (re-uploading) is common for videos.
However, the principle of tackling multi-platform distribution gaps remains unclear.

\textit{2) Multi-domain detection.} Misinformation texts in different news domains have different word use and propagation patterns, leading to data shifts~\cite{mdfend,zhu2022memory,sheng2022characterizing}. Therefore, the investigation and mitigation of the domain gap for video misinformation is a promising direction. 

\textit{3) Temporal generalization.} Distribution shift over time is unavoidable for online video data. Effective features on past data might perform poorly online~\cite{zhu2022generalizing,mu-etal-2023-time,hu-etal-2023-learn}. How to find stable features and rapidly adapt to current video data require further exploration.

\subsubsection{Explainability}
Most existing methods focus on improving accuracy and neglect the importance of providing an explanation. Without explanations aligned with human expectations, human users could hardly learn about the strengths and weaknesses of a detection system and decide when to trust it. This issue should be tackled in two aspects:

\textit{1) Distinguishing fine-grained types of misinformation.} In addition to binary classification, the model should further predict a concrete type of detected misinformation samples (e.g., video misuse). This requires a new taxonomy and fine-grained annotation on datasets.

\textit{2) Multi-modal clue attribution.} The detection model should attribute its output to the multi-modal clues extracted. Due to the complicated characteristics and fine-grained types of misinformation videos, a single piece of misinformation video, on the one hand, may only have a few clues. On the other hand, normal real videos can also contain some aforementioned features, e.g. editing traces, and propaganda intentions. It is important to disentangle the clue integration process to justify the final output by clue attribution, whether the video is real or fake.

\subsection{Future Directions}
\subsubsection{Clue Integration \& Reasoning}
The diversity of involved modalities in a video post requires the detection model to have a higher clue integration and reasoning ability than that for text- and image-based detection. In most cases, the final judgment of misinformation depends on neither a single modality nor all modalities, and finding out effective combinations is non-trivial. 
For example, for a previously recorded accident video that is repurposed to be the scene of a new accident and added background music, what is crucial for judgment is the mismatch between video and text, rather than that between video and audio.

However, clue integration in this area is typically accomplished by directly aligning and fusing all representation vectors obtained from different modalities, which makes it hard for models to learn to reason among modalities. We believe that enabling reasoning among modalities will be important for better clue integration and more flexible detection. The possible directions include: 

\textit{1) Inter-modality relationship modeling.} Following tasks requiring reasoning ability like visual question answering~\cite{gao2020multi}, one can build graphs to guide interaction among modalities.

\textit{2) Problem decomposition.} By transforming the detection as a mixture of several subproblems, one can use Chain/Tree of Thoughts~\cite{wei2022chain,yao2023tree} to prompt large language models (e.g., GPT-4~\cite{gpt4})
to reason. This is useful for video claim verification when combined with external knowledge sources like online encyclopedias, fact-checking reports and relevant news (similar to that in text and image format~\cite{thorne2018fever,sheng-etal-2021-article,abdelnabi2022open,sheng-etal-2022-zoom}).

\subsubsection{Recommendation-Detection Coordination}

The coordination between recommendation-based video distribution and misinformation video detection is crucial for practical systems, whose ultimate goal is to keep recommending videos that are of interest to users while avoiding misinforming them. To achieve this, detection systems are expected to contain different models and strategies to exploit rich side information from recommender systems as well as make recommendations more credible.
Here we provide three concrete coordination scenarios: 

\textit{1) User-interest-aware detection.} The viewing history of the videos reflects not only users' interests but also how susceptible they are to specific topics (e.g., elections). Therefore, we could prioritize these recommended videos and detect misinformation with awareness of topics (a similar case for text fake news is~\cite{wang2022veracity}).

\textit{2) User-feedback-aware detection.} Feedback from the crowd to the platform might be valuable indicators of suspicious videos. A recent example is to use users' reports of misinformation as weak supervision in text-based fake news detection~\cite{wang2020weak}. Using more user feedback derived from recommender systems like expressions of dislike due to factuality issues will be a promising direction.

\textit{3) Credibility-aware recommendation.} Considering information credibility in recommender systems can mitigate the exposure of misinformation videos and make the recommendation more accountable. A possible solution is to include misinformation video detection as an auxiliary task or use a well-trained detector as a critic to provide feedback. 

\section{Conclusion}
\label{sec:conclu}
We surveyed the existing literature on misinformation video detection and provided an extensive review of the advanced detection solutions, including clues at the signal, semantic, and intent levels and clue integration techniques. We also summarized publicly available datasets and tools and discussed related areas. Furthermore, we presented critical open issues and future directions for both research and real-world applications. 
Also, we open-sourced a GitHub repository that will be updated to include future advances in this area.
We hope this survey could shed light on further research for defending against misinformation videos.

\begin{acks}
This research is supported in part by the National Key Research and Development Program of China (2022YFC3302102), the National Natural Science Foundation of China (62203425), the Project of Chinese Academy of Sciences (E141020), the China Postdoctoral Science Foundation (2022TQ0344), and the International Postdoctoral Exchange Fellowship Program by Office of China Postdoc Council (YJ20220198).
\end{acks}

\bibliographystyle{ACM-Reference-Format}
\balance
\bibliography{VideoSurvey}

\appendix
\begin{figure*}
\centering
\begin{subfigure}{0.3\textwidth}
    \includegraphics[width=\textwidth]{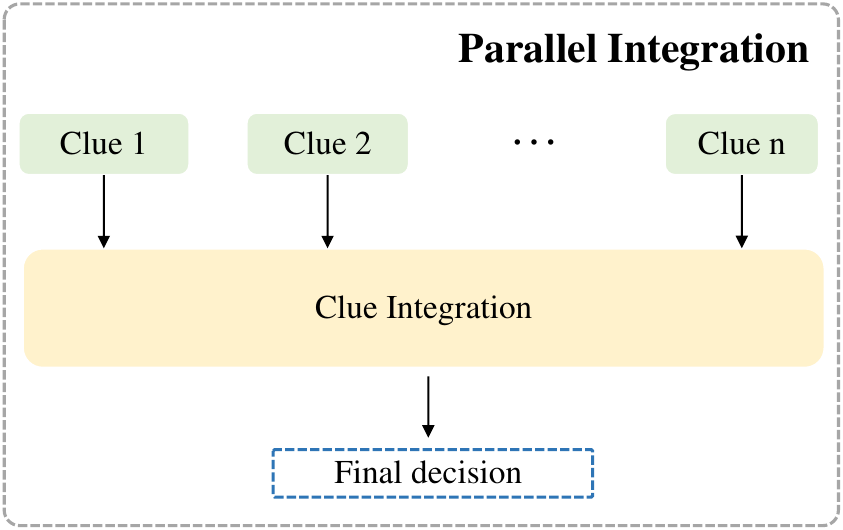}
\end{subfigure}
\hspace{0.05\textwidth}
\begin{subfigure}{0.3\textwidth}
    \includegraphics[width=\textwidth]{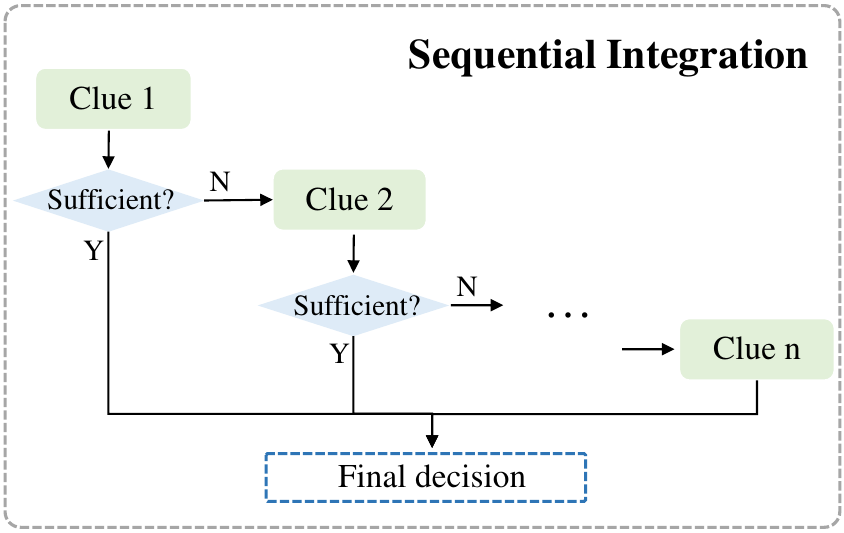}
\end{subfigure}
\caption{Comparison of the parallel and sequential clue integration in misinformation video detection.}
\label{fig:appendix_flowchart}
\end{figure*}

\begin{figure*}
\centering
\label{fig:appendix_models}
\begin{subfigure}{0.224\textwidth}
    \includegraphics[width=\textwidth]{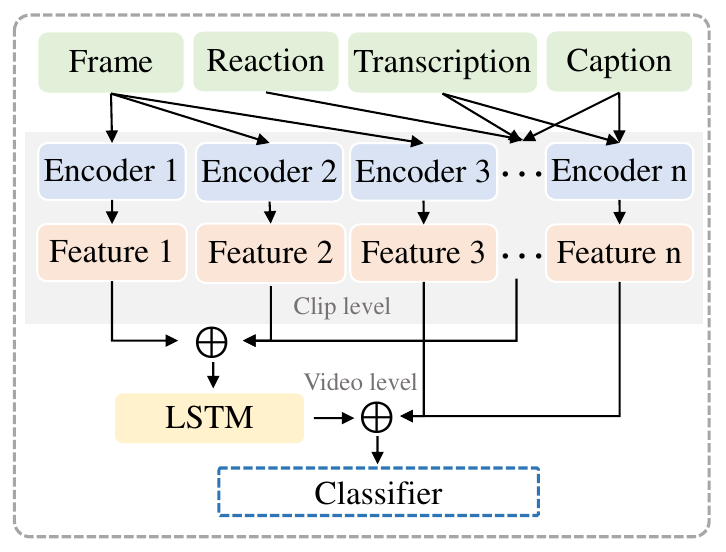}
    \caption{\citet{mccrae2022multi}}
    \label{fig:mccrae}
\end{subfigure}
\hspace{0.01\textwidth}
\begin{subfigure}{0.22\textwidth}
    \includegraphics[width=\textwidth]{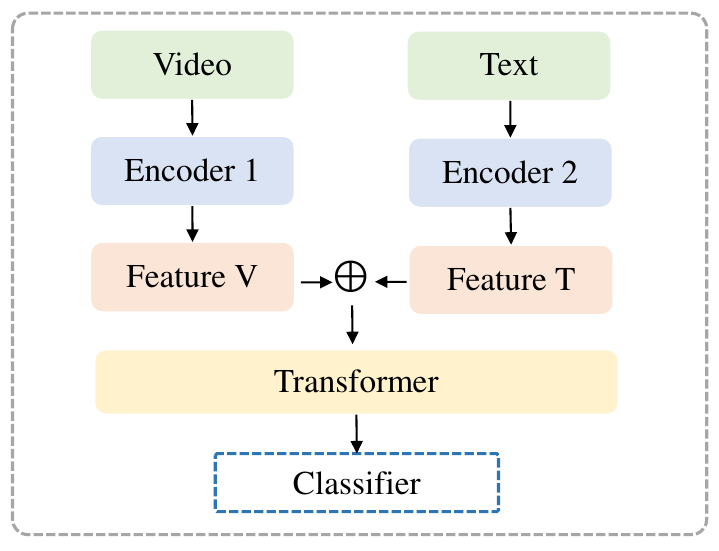}
    \caption{\citet{wang2022misinformation} (MLM)}
    \label{fig:wang_mlm}
\end{subfigure}
\hspace{0.01\textwidth}
\begin{subfigure}{0.22\textwidth}
    \includegraphics[width=\textwidth]{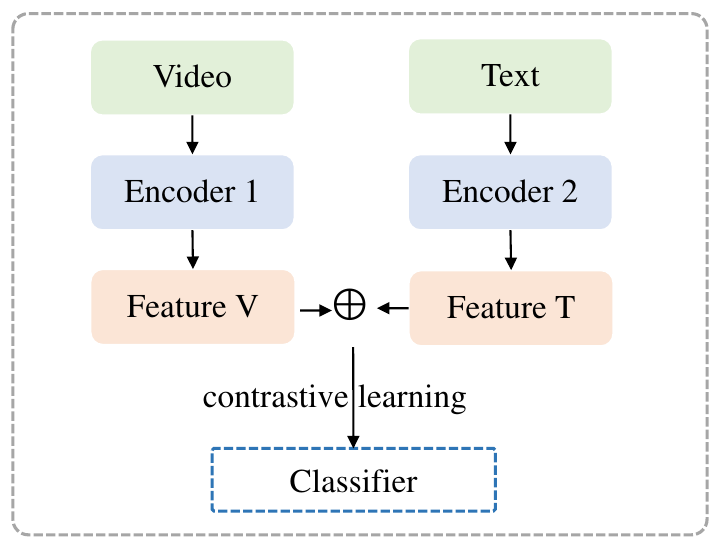}
    \caption{\citet{wang2022misinformation} (CL)}
    \label{fig:wang_cl}
\end{subfigure}
\hspace{0.01\textwidth}
\begin{subfigure}{0.22\textwidth}
    \includegraphics[width=\textwidth]{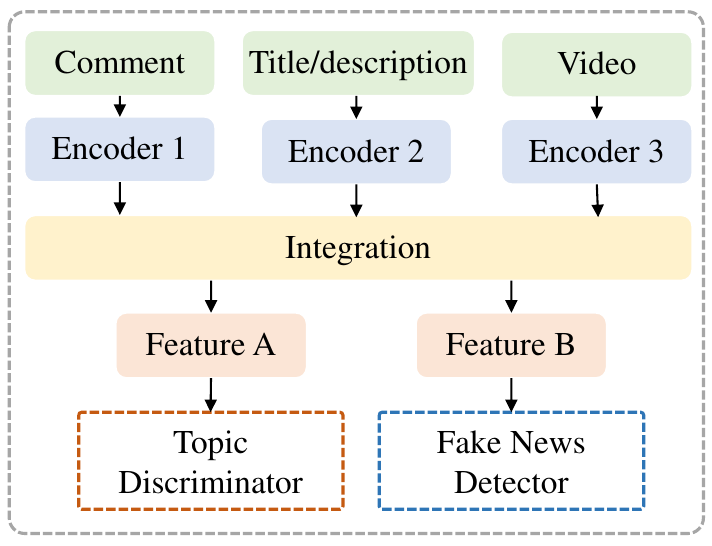}
    \caption{\citet{choi2021cikm}}
    \label{fig:choi}
\end{subfigure}

\caption{Illustrations of representative misinformation video detection models.}
\end{figure*}

\begin{table*}[h]
\centering
\caption{Overview of datasets for misinformation video detection. Note that \citet{christodoulou2023identifying} conducted experiments on a subset of YouTubeAudit. SOTA: state-of-the-art. \label{tab:dataset_overview}}
\resizebox{0.95\textwidth}{!}{
\begin{tabular}{cccrcrcc} 
\toprule
\multirow{2}{*}{\textbf{Dataset}} & \multirow{2}{*}{\textbf{Sources}}      & \multirow{2}{*}{\textbf{Domain}} & \multirow{2}{*}{\textbf{Size}} & \multirow{2}{*}{\textbf{Released?}} & \multirow{2}{*}{\textbf{SOTA Method}} & \multicolumn{2}{c}{\textbf{SOTA Performance (\%)}}  \\
                         &                              &                         & \multicolumn{1}{l}{}                        &                            &                              & \textbf{Accuracy} & \textbf{F1}                              \\ 
\midrule
FVC                   & YouTube, Twitter, Facebook~~ & various                & 5,006   & Y         & FANVM~\cite{choi2021cikm}                    & ——       & 85.84                           \\
VAVD                  & YouTube                      & various                & 546     & Y         & FANVM~\cite{choi2021cikm}                    & ——       & 86.71                           \\
YouTubeAudit          & YouTube                      & Conspiracy theories & 2,943   & Y         & (subset) RoBERTa~\cite{christodoulou2023identifying}          & 94.00    & 94.00                           \\
FakeSV                & Douyin, Kuaishou             & various                & 5,538   & Y         & SVFND+NEED~\cite{qi2023two}               & 84.62    & 84.61                           \\
COVID-VTS             & Twitter                      & COVID-19            & 10,000  & Y         & TwtrDetective~\cite{liu2023covid}            & 68.10    & 67.90                           \\ 
\midrule
(Hou et al. 2019)     & YouTube                      & Prostate cancer     & 250     & N         & (Hou et al.)~\cite{hou2019icmi}             & 74.41    & ——                              \\
(Serreno et al. 2020) & YouTube                      & COVID-19            & 180     & N         & (Serreno et al.)~\cite{aclworkshop2020}         & 84.40    & ——                              \\
(Jagtap et al. 2021)  & YouTube                      & various                & 2,125   & N         & (Jagtap et al.)~\cite{jagtap2021misinformation}          & 92.00    & 78.00                           \\
(Shang et al. 2021)   & YouTube                      & COVID-19            & 891     & N         & TikTec~\cite{shang2021tiktec}                  & 72.31    & 60.51                           \\
(Choi and Ko 2021)    & YouTube                      & various                & 4,622   & N         & FANVM~\cite{choi2021cikm}                    & ——       & 86.28                           \\
(Choi and Ko 2022)    & YouTube                      & various                & 2,912   & N         & FVDM~\cite{choi2022prl}                     & ——       & 82.55                           \\
(Li et al. 2022)      & Bilibili                     & Health              & 700     & N         & (Li et al.)~\cite{li2022cnn}              & 90.00    & 89.00                           \\
(Wang et al. 2022a)   & Twitter                      & various                & 160,000 & N         & (Wang et al.)~\cite{wang2022misinformation}            & 71.07    & ——                              \\
(Wang et al. 2022b)   & Twitter                      & various                & 943,667 & N         & (Wang et al.)~\cite{wang2022msmd}            & 85.43    & ——                              \\
(McCrae et al. 2022)  & Facebook, YouTube             & various                & 4,651   & N         & (McCrae et al. )~\cite{mccrae2022multi}         & 60.50     & ——                              \\
(Yang et al. 2023)    & Douyin, Kuaishou             & various                & 8,213   & N         & (Yang et al)~\cite{yang2023multimodal}             & 87.71    & 87.38                           \\
\bottomrule
\end{tabular}}
\end{table*}

\section{Additional Illustrations}
For better clarification, we visualize and compare the processes of the two paradigms of clue integration, as presented in Fig.~\ref{fig:appendix_flowchart}. Since most existing works adopt parallel integration, here we present four representative models to help readers gain an intuitive understanding of detection frameworks. Figs.~\ref{fig:mccrae} and~\ref{fig:wang_mlm} respectively demonstrate the concatenation-based fusion and the utilization of attention mechanisms to model cross-modal interactions. Figs.~\ref{fig:wang_cl} and~\ref{fig:choi} respectively showcase the employment of auxiliary objectives like contrastive learning and topic classification to optimize the learning of multi-modal features.

\section{Overview of Datasets}
\label{sec:appendix_dataset}
In Sec.~\ref{sec:dataset}, we delve into the detailed description of four large publicly available datasets. However, it is worth noting that there are several small-scale and topic-specific datasets that have not been open-sourced but still contribute significantly to the research. Table~\ref{tab:dataset_overview} summarizes all these datasets, including their data source, domain, size, accessibility, state-of-the-art methods, and the corresponding performance based on accuracy and F1 scores.
\end{document}